\documentclass[sigconf]{acmart}
\renewcommand\footnotetextcopyrightpermission[1]{} 

\usepackage{fancyhdr}
\makeatletter
\@ifundefined{Bbbk}{}{}
\makeatother

\usepackage{graphicx}
\usepackage{amsmath}
\usepackage{hyperref}
\usepackage{xcolor}
\usepackage{tcolorbox}
\tcbuselibrary{listingsutf8}
\usepackage[skip=2pt]{caption}

\graphicspath{{./images/}{./appendix_images/}}

\begin{document}

\title{Deciphering the Underserved: Benchmarking LLM OCR for Low-Resource Scripts}

\author{Muhammad Abdullah Sohail}
\email{26100142@lums.edu.pk}
\affiliation{%
  \institution{Lahore University of Management Sciences}
  \city{}
  \state{}
  \country{}
}

\author{Salaar Masood}
\email{26100149@lums.edu.pk}
\affiliation{%
  \institution{Lahore University of Management Sciences}
  \city{}
  \state{}
  \country{}
}

\author{Hamza Iqbal}
\email{26100130@lums.edu.pk}
\affiliation{%
  \institution{Lahore University of Management Sciences}
  \city{}
  \country{}
}

\begin{abstract}
This study investigates the potential of Large Language Models (LLMs), particularly GPT-4o, for Optical Character Recognition (OCR) in low-resource scripts such as Urdu, Albanian, and Tajik, with English serving as a benchmark. Using a meticulously curated dataset of 2,520 images incorporating controlled variations in text length, font size, background color, and blur, the research simulates diverse real-world challenges. Results emphasize the limitations of zero-shot LLM-based OCR, particularly for linguistically complex scripts, highlighting the need for annotated datasets and fine-tuned models. This work underscores the urgency of addressing accessibility gaps in text digitization, paving the way for inclusive and robust OCR solutions for underserved languages.

\end{abstract} 

\maketitle
\section{Introduction}
Optical Character Recognition (OCR) has become an indispensable technology for digitizing textual content, converting written information into machine-readable formats, and enabling broad accessibility. Despite decades of advances, OCR research has primarily centered on resource-rich languages and well-studied scripts, leaving many others, especially those with complex structural features such as intricate ligatures and contextual dependencies, unexplored \cite{mustafa2024parseq, sarkar2024lowresource}. These underserved languages, often referred to as "low-resource languages," lack sufficient annotated datasets, linguistic tools, or computational resources for effective natural language processing (NLP) tasks \cite{joshi2020state, nekoto2020participatory}.

In recent years, LLMs have demonstrated transformative potential across a range of NLP tasks. These models excel at processing and generating text across diverse contexts, offering a dynamic, context-aware alternative to conventional OCR systems. Unlike traditional approaches that rely heavily on predefined patterns and annotated datasets, LLMs leverage their multimodal capabilities to adapt to diverse linguistic and structural complexities \cite{zhu2023llmocr}. However, their application to OCR remains nascent, particularly for scripts that present both linguistic and visual challenges \cite{2404.02375v1}.

\section{Related Works}

Significant advancements in OCR, particularly for high-resource scripts, have transitioned from traditional rule-based systems to machine learning and deep learning approaches. Early OCR systems relied on segmentation-based pipelines, incorporating methods such as template matching and Support Vector Machines (SVMs) for feature classification \cite{lehal2000gurmukhi, pal1998complete}. However, these approaches relied heavily on language-specific heuristics, limiting their generalizability to complex scripts \cite{bashir2022kashmiri}. Neural network-based OCR models redefined the field, enabling sequence modeling and direct text-to-label transcription using architectures like Bidirectional Long-Short Term Memory (BLSTM) with Connectionist Temporal Classiﬁcation (CTC) loss \cite{krishnan2014robust, sankaran2013transcription}. Synthetic datasets further enhanced these models by mitigating the scarcity of annotated data for specific languages \cite{jaderberg2014synthetic, sarkar2024lowresource}

Progress in low-resource OCR remains limited, with studies focusing on Indic scripts, while structurally complex scripts—such as those with ligatures or modified Cyrillic alphabets—remain underrepresented \cite{sarkar2024lowresource}. Similarly, few studies have explored OCR pipelines for Urdu and Bengali \cite{hakro2016sindhi, bashir2022kashmiri}, highlighting the need for customized solutions. LLMs have recently emerged as a promising OCR solution, leveraging multimodal capabilities for dynamic adaptation to diverse scripts and challenging visual conditions \cite{zhu2023llmocr, 2404.02375v1}. Studies have demonstrated LLMs' ability to process textual content from images, outperforming traditional OCR models in certain contexts \cite{mustafa2024parseq}. Nevertheless, their application to low-resource scripts under controlled conditions remains underexplored. This highlights the urgent need for focused research to leverage LLMs in bridging these gaps and advancing the digitization of underserved, complex scripts.

\section{Dataset Curation}

For our study, we meticulously constructed a high-quality benchmark dataset consisting of 2,520 images to evaluate the OCR performance of LLMs. This approach incorporated various dimensions across linguistics and visual conditions.

\subsection{Language Selection and Sources}
Our approach included curating a dataset across four languages—Urdu, English, Albanian, and Tajik—carefully chosen to ensure both linguistic diversity and relevance to the study's objectives. Notably, each of these languages holds the status of a national language in their respective countries—Urdu in Pakistan, English globally including its official role in numerous nations, Albanian in Albania, and Tajik in Tajikistan. This distinction underscores their cultural and linguistic richness, yet these languages remain low-resource in natural language processing (NLP) and OCR research. To this end, our study highlights the transformative potential of OCR systems in addressing critical accessibility and digitization gaps

\subsubsection{English}
English was included as a benchmark language owing to its status as a high-resource language with extensive datasets and tools already available in the NLP domain. The English dataset was derived by scraping news from Samaa TV and Geo News, resulting in a total of 1,288 articles. 

\subsubsection{Urdu}
Urdu was selected for its intricate Nastaliq script, characterized by ligatures, bidirectional text, and diacritical marks, which pose significant OCR challenges. Articles were sourced from Urdu news outlets including Jang, Samaa TV, Geo News, and Express News, yielding 2,043 entries.

\subsubsection{Albanian}
Albanian has a distinct linguistic structure, including diacritic-rich text. However, its script is quite similar to that of English, which allows for an interesting evaluation of OCR performance across scripts with shared Latin roots. Articles for the Albanian dataset were sourced from Koha Jonë, a leading Albanian news outlet, resulting in 1,108 articles.

\subsubsection{Tajik}
Tajik is a language written in a modified Cyrillic alphabet. Articles for this language were collected from Khovar, the national news agency of Tajikistan, yielding a total of 1,050 articles.


\subsection{Design and Augmentation}
The dataset design incorporated controlled visual variations across four key dimensions: word count ranges, font sizes, blur and background colours.

We defined three ranges of word count, namely 40-60, 110-130, and 180-200. A total of 30 articles per language were selected for each word count range. This selection resulted in 90 articles (30 articles per word count range × 3 ranges). The text from these articles was manually formatted into images using PowerPoint. 

Each article was rendered in three font sizes: 12, 18, and 24 points. These font sizes were selected to reflect common variations in printed and digital text. Size 12 simulated compact text typically found in dense layouts, size 18 served as a balanced baseline that was neither too small nor too large, and size 24 represented larger, more readable text often used for emphasis or headlines. Screenshots of the formatted slides were taken to generate 90 images per word count range (30 articles × 3 font sizes). This process was extended for the other two ranges of word count, resulting in 270 images.

To further simulate real-world OCR challenges, we introduced two augmentation dimensions—background color and Gaussian blur. The background color augmentations included \textit{Slate Gray} (\#708090), representing low-contrast conditions, and \textit{Light Yellow} (\#FFFACD), simulating high-contrast but visually distinct settings, in addition to the base \textit{White} (\#FFFFFF), which was kept as a control. These variations tested the robustness of the OCR against different text-to-background contrast levels, a common challenge in document digitization tasks. Similarly, Gaussian blur augmentations at levels 0.75 and 1.5 were added to simulate motion blur and slight defocus, while 0 (no blur) served as the baseline, reflecting clear, undistorted images.

These augmentations were applied exclusively to the base images formatted at font size 18, chosen as the baseline due to its balanced legibility and challenge. Each augmentation dimension generated 60 additional images per word count range (30 base images × 2 variations), resulting in 120 augmented images. This brought the total to 210 images for each word count range. Across the three word count ranges, this augmentation yielded 630 images per language (210 images × 3 ranges).

\subsection{Final Composition}
The final dataset comprises 2,520 images, evenly distributed across the four languages—Urdu, English, Albanian, and Tajik—with 630 images per language. Each language dataset incorporates controlled variations in word count, font size, background color, and blur levels, ensuring a comprehensive representation of linguistic and visual challenges for OCR evaluation. Sample images from the dataset are also provided in the Appendix section of this paper. The detailed composition of the dataset across these dimensions is summarized in Table~\ref{tab:dataset_composition}.

\vspace{-0.4em}
\begin{table}[h!]
\centering
\scriptsize
\caption{Dataset Composition Across Each Language} 
\vspace{-0.4em}
\label{tab:dataset_composition}
\begin{tabular}{lccccccc}
\hline
\textbf{Dimension} & \textbf{Font} & \textbf{Blur} & \textbf{BG Color} & \textbf{40-60} & \textbf{110-130} & \textbf{180-200} & \textbf{Total} \\ \hline
Word Count         & 18            & 0             & White             & 30             & 30               & 30               & 90             \\
Font Size          & 12            & 0             & White             & 30             & 30               & 30               & 90             \\
                   & 24            & 0             & White             & 30             & 30               & 30               & 90             \\
Blur               & 18            & 0.75          & White             & 30             & 30               & 30               & 90             \\
                   & 18            & 1.5           & White             & 30             & 30               & 30               & 90             \\
BG Color           & 18            & 0             & Slate Gray        & 30             & 30               & 30               & 90             \\
                   & 18            & 0             & Light Yellow      & 30             & 30               & 30               & 90             \\ \hline
\textbf{Total}     &               &               &                   &                &                  &                  & \textbf{630}   \\ \hline

\end{tabular}
\end{table}

\vspace{-1.2em}
\section{Methodology}

This study adopts a structured experimental framework, combining controlled conditions and rigorous evaluation metrics to systematically assess the OCR performance of LLMs across diverse linguistic and visual challenges. 


\subsection{Experimental Setup}

To evaluate the performance of LLM-based OCR systems, we employed GPT-4o, a state-of-the-art multimodal large language model. The model was operated in a zero-shot inference mode, leveraging its advanced contextual understanding to transcribe text images without prior fine-tuning. Images were encoded in Base64 format before being processed by the model. A carefully crafted system prompt ensured precise and consistent outputs by instructing the model to focus solely on text extraction while preserving linguistic fidelity. The system prompt was structured as follows:

\begin{quote}
\texttt{"As an AI language model specialized in Optical Character Recognition (OCR) for \{\textit{language}\} script, your task is to extract the text from the provided image accurately.\\\\
You are required to follow the following instructions:\\\\
- Output only the text found in the image.\\
- Do not add any additional commentary or information.\\
- Preserve all diacritical marks and nuances of the script\\
- Do not translate the text."}
\end{quote}

To ensure reproducibility, the inference pipeline was fully automated with consistent preprocessing steps and the process was executed with a temperature setting of \texttt{temp=0} to ensure deterministic outputs, a critical requirement for controlled experiments. The results were systematically analyzed to identify patterns and thresholds, providing actionable insights into the adaptability of LLM-based OCR systems under diverse linguistic and visual challenges.


\subsection{Evaluation Metrics}

Performance evaluation employed three widely used metrics in OCR research: Character Error Rate (CER), Word Error Rate (WER), and BLEU score. These metrics collectively provide a comprehensive assessment of accuracy and linguistic quality, each offering unique insights into the system's performance.

\subsubsection{Character Error Rate (CER)}
CER measures character-level accuracy by quantifying the edit distance between the predicted text and the reference text. It accounts for the minimum number of character substitutions ($S$), insertions ($I$), and deletions ($D$) required to transform the predicted text into the reference text. CER is defined as:

\vspace{-1em}
\begin{equation}
\text{CER} = \frac{S + I + D}{N},
\end{equation}
where $N$ is the total number of characters in the reference text. In the context of our project, CER serves as a crucial metric, particularly for scripts like Urdu and Tajik, where diacritic richness, complex ligatures, and character-level intricacies demand precise recognition.

\subsubsection{Word Error Rate (WER)}
WER evaluates word-level fidelity by extending CER to entire words. It captures the number of word-level substitutions, insertions, and deletions required to align the predicted text with the reference text. WER is calculated as:

\vspace{-0.5em}
\begin{equation}
\text{WER} = \frac{S + I + D}{W},
\end{equation}
where $W$ is the total number of words in the reference text. WER plays a critical role in our study by evaluating word-level recognition accuracy under varying dimensions, offering insights into how these factors impact the preservation of semantic meaning across languages, especially for low-resource scripts where word-level errors can significantly degrade overall readability.

\subsubsection{BLEU Score}
BLEU Score (Bilingual Evaluation Understudy) assesses the fluency and semantic coherence of the predicted text by comparing its $n$-grams with those of the reference text. Originally developed for machine translation, BLEU is adapted to OCR to measure the quality of transcriptions. It is defined as:

\vspace{-0.5em}
\begin{equation}
\text{BLEU} = BP \cdot \exp\left( \sum_{n=1}^{N} w_n \log p_n \right),
\end{equation}
where $p_n$ is the precision of $n$-grams, $w_n$ is the weight assigned to each $n$-gram, and $BP$ is the brevity penalty applied to shorter predicted outputs. For our study, BLEU serves as a vital metric to evaluate the semantic and fluency-oriented impact of various dimensions on OCR performance, particularly highlighting how well the transcriptions preserve the linguistic coherence of low-resource scripts under differing conditions.

\section{Results and Discussion}

Our study examines the influence of four dimensions, namely word count, font size, background color, and Gaussian blur. To ensure precise evaluation and eliminate confounding effects, baseline settings were defined: no blur, a white background, font size 18, and a word count range of 110–130. While assessing each dimension, only its parameters were varied, with all others held constant at the set baseline conditions. This rigorous approach isolates the impact of individual factors, providing a clear understanding of their interaction with the structural and visual complexities of the scripts under study.

\subsection{Impact of Word Count}

\begin{figure}[!htb]
    \centering
    \vspace{-0.57em} 
    \includegraphics[width=0.48\textwidth, height=0.4\textwidth, keepaspectratio]{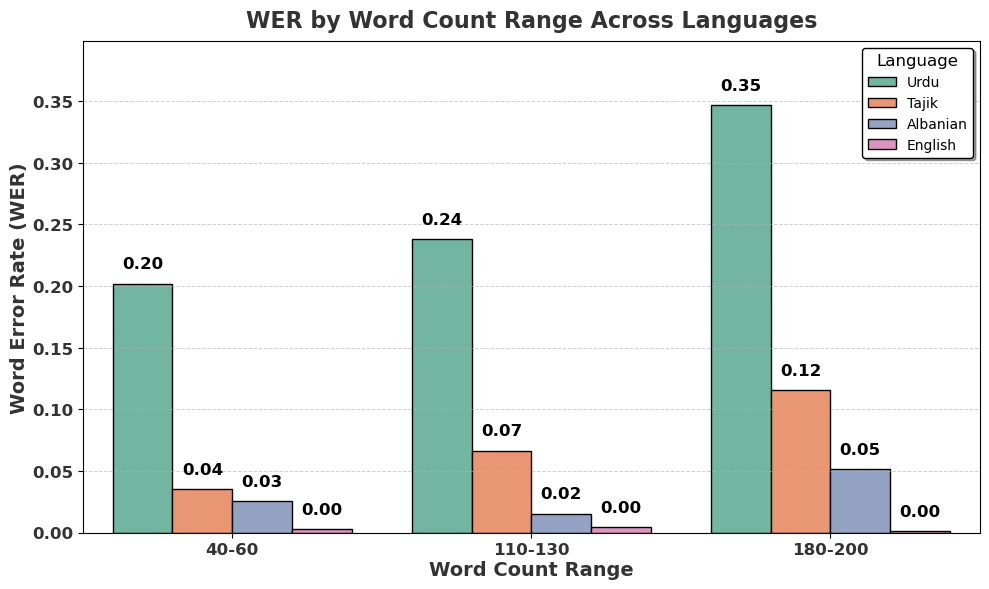}
    \vspace{-1.5em} 
    \caption{\small WER with Increasing Word Count}
    \label{fig:wer_word_count}
    \vspace{-0.6em} 
\end{figure}

Our analysis highlights a clear divergence across scripts on varying word count ranges, with performance degradation being most pronounced in complex, low-resource languages such as Urdu. Figure~\ref{fig:wer_word_count} shows Urdu’s WER rising sharply from 0.20 for shorter texts (40–60 words) to 0.35 for longer ones (180–200 words), exposing the challenges of its intricate ligatures, bidirectional structure, and diacritical density. As the text length increases, error accumulation compounds, reflecting the model’s limitations when handling complex sequences. For Tajik, WER rises more gradually (0.04 to 0.12), indicating moderate sensitivity due to its simpler Cyrillic script. In contrast, Albanian and English remain relatively stable across word ranges, with near-zero WER for English, owing to their structural simplicity and extensive representation of Latin scripts in GPT-4o’s training data.

\vspace{-0.4em}

\begin{figure}[h!]
    \centering
    \includegraphics[width=0.48\textwidth, height=0.4\textwidth, keepaspectratio]{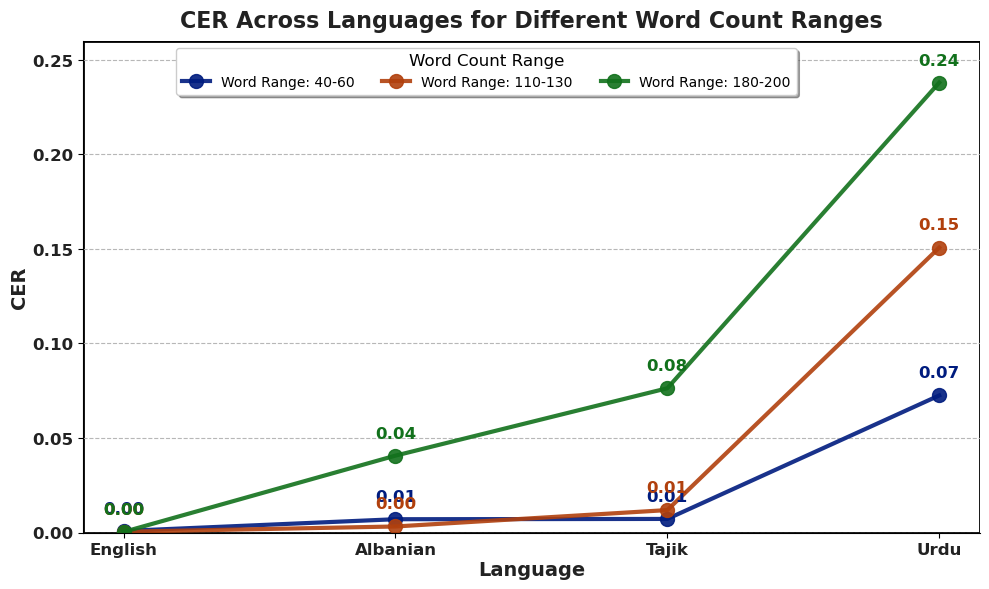}
    \caption{\small CER with Increasing Word Count}
    \label{fig:cer_word_count}
\end{figure}

A similar pattern emerges in Figure~\ref{fig:cer_word_count} for CER, where Urdu’s character-level accuracy declines sharply from 0.07 to 0.24 as word count increases, reflecting the challenge of aligning extended sequences in its dense script. Tajik shows a moderate rise, while Albanian and English remain stable. These results emphasize the need for greater training exposure to low-resource scripts like Urdu and Tajik, including annotated datasets.

\vspace{-0.8em}
\vspace{-0.6em}

\begin{figure}[h!]
    \centering
    \includegraphics[width=0.48\textwidth, height=0.4\textwidth, keepaspectratio]{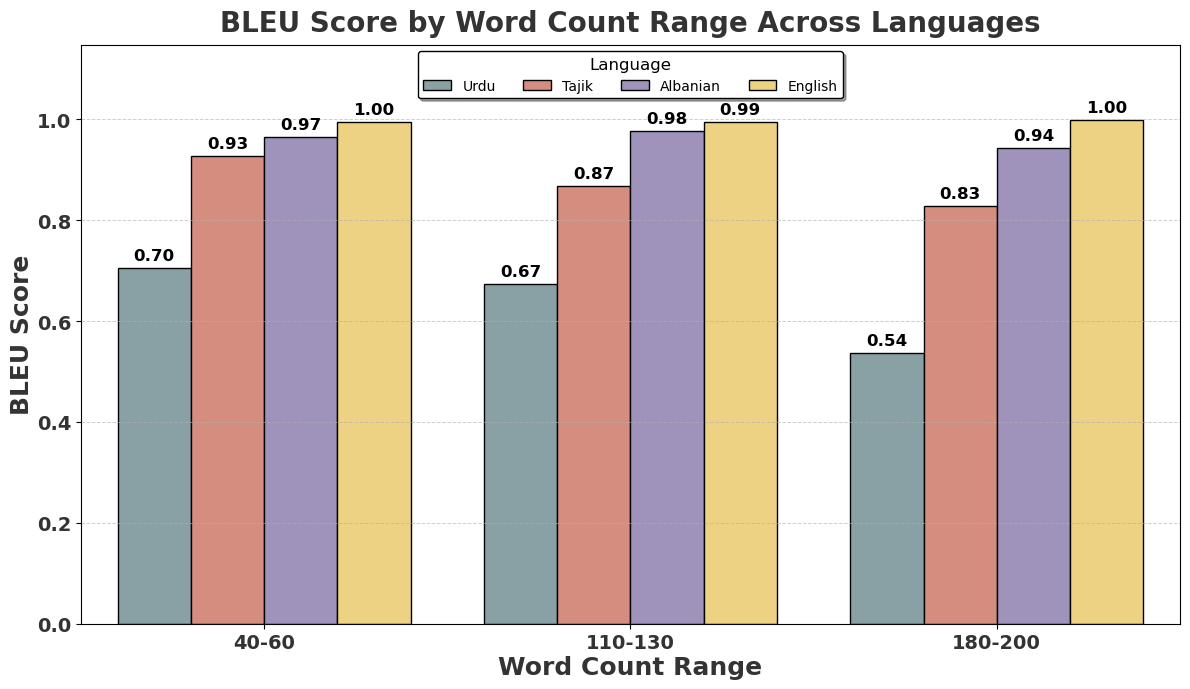}
    \caption{\small BLEU score with Increasing Word Count}
    \label{fig:bleu_word_count}
\end{figure}

Figure~\ref{fig:bleu_word_count} highlights the decline in BLEU scores, with Urdu dropping from 0.70 to 0.54 as word count increases, reflecting how recognition errors distort semantic coherence in its diacritic-heavy script. Tajik shows a milder decline, while Albanian and English remain stable with near-perfect scores. This trend underscores the model’s struggle with complex, underrepresented scripts as sequence length grows. To address this, expanding annotated datasets and implementing script-aware techniques like ligature segmentation are crucial for improving OCR accuracy in low-resource languages.

\vspace{-0.5em}
\vspace{-1.5em}

\subsection{Impact of Font Size}

\begin{figure}[h!]
    \centering
    \includegraphics[width=0.48\textwidth, height=0.4\textwidth, keepaspectratio]{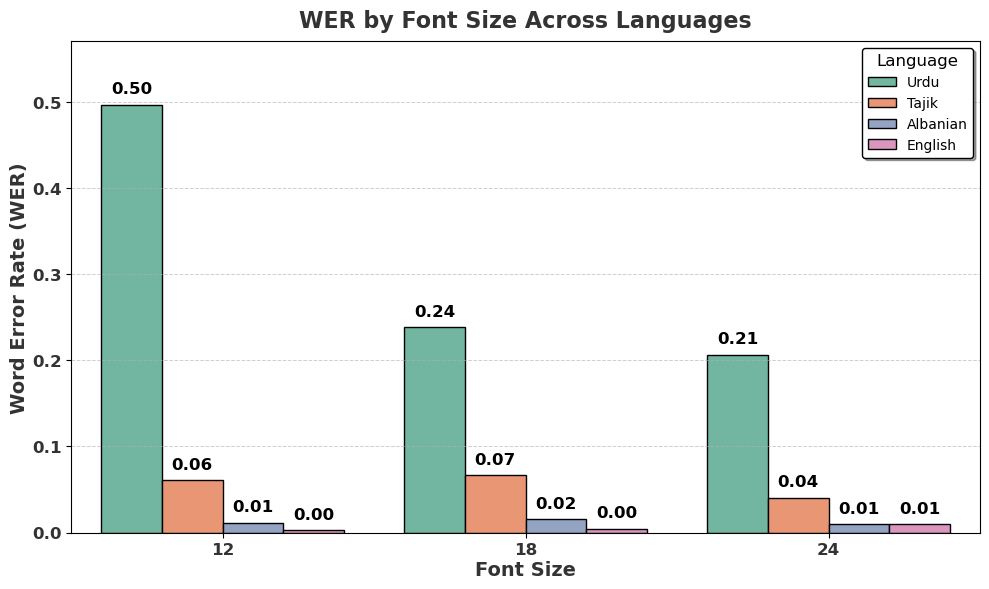}
    \caption{\small WER with Increasing Font Size}
    \label{fig:wer_font_size}
\end{figure}

Font size significantly influences OCR performance, with smaller sizes amplifying errors due to reduced visual clarity, particularly for complex scripts. Figure~\ref{fig:wer_font_size} shows Urdu’s WER dropping sharply from 0.50 at size 12 to 0.24 at 18 and 0.21 at 24, highlighting the challenge of dense ligatures and diacritics merging at smaller scales. For Tajik, WER decreases from 0.06 to 0.04 as the font size increases, reflecting its comparatively simpler Cyrillic structure but still indicating sensitivity to smaller scales. In contrast, Albanian and English maintain consistently low WER across all font sizes, benefiting from their linear script structures and extensive representation in GPT-4o’s pre-training data, which reinforces their resilience to visual compression.

\vspace{-1em}
\vspace{-1em}

\begin{figure}[h!]
    \centering
    \includegraphics[width=0.48\textwidth, height=0.4\textwidth, keepaspectratio]{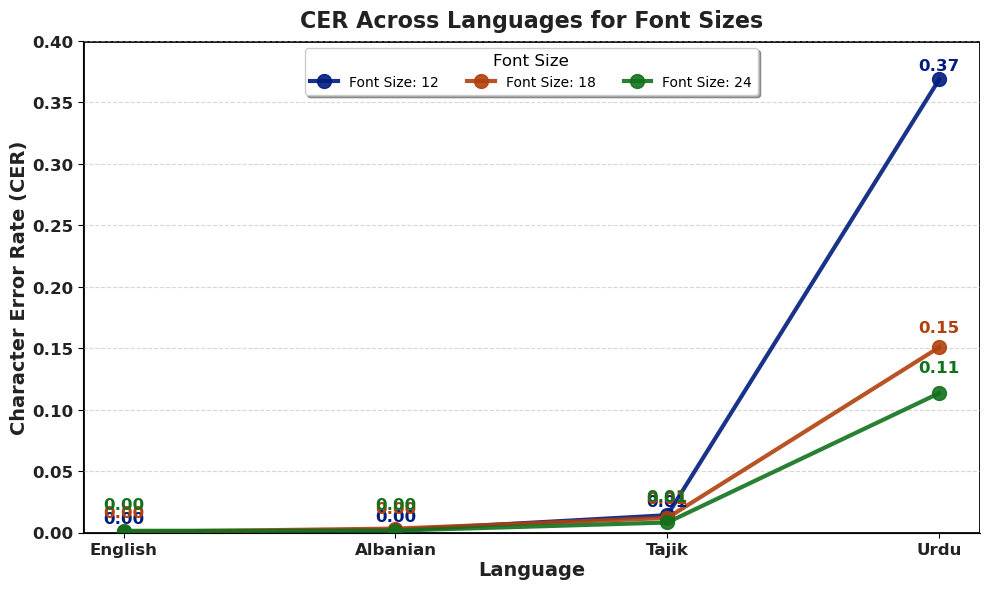}
    \caption{\small CER with Increasing Font Size}
    \label{fig:cer_font_size}
\end{figure}

\begin{figure}[h!]
    \centering
    \includegraphics[width=0.48\textwidth, height=0.4\textwidth, keepaspectratio]{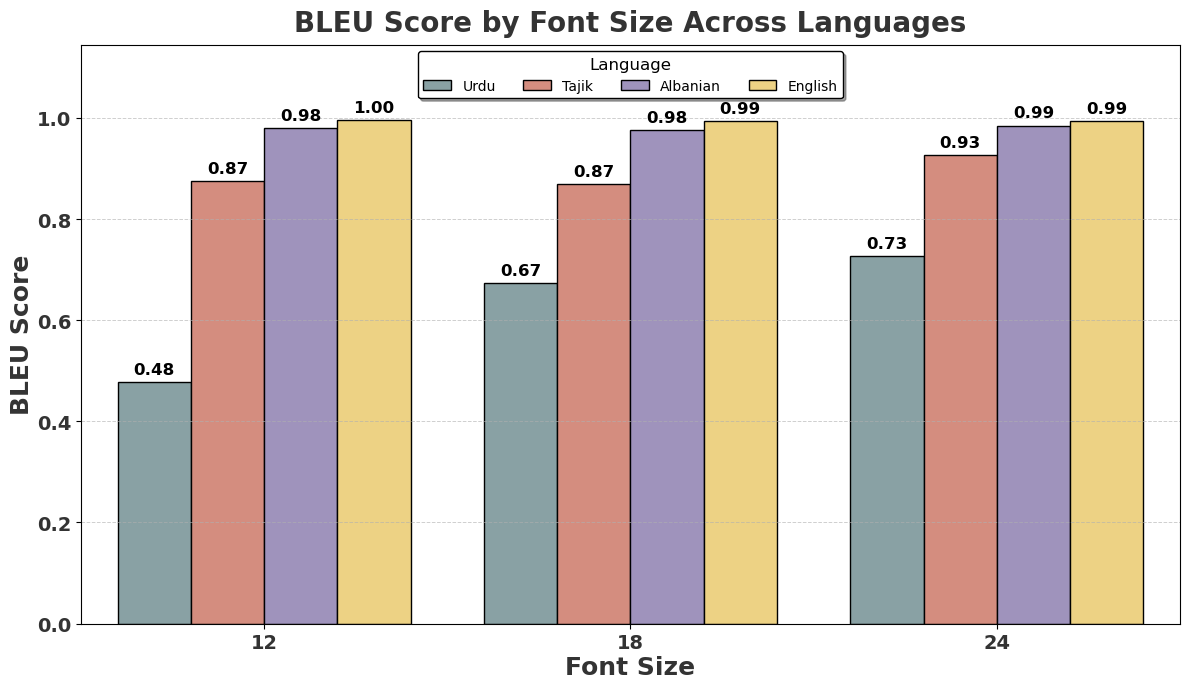}
    \caption{\small BLEU score with Increasing Font Size}
    \label{fig:bleu_font_size}
\end{figure}

Character-level performance, shown in Figure~\ref{fig:cer_font_size}, follows a similar trend, with CER for Urdu declining steeply as font size increases, while Tajik exhibits moderate improvement. Smaller font sizes exacerbate alignment errors in scripts with dense visual details, where precise diacritical recognition is critical. Albanian and English remain unaffected, with CER remaining minimal across all sizes. Figure~\ref{fig:bleu_font_size} reveals the semantic impact of these trends, where BLEU scores for Urdu rise from 0.48 at size 12 to 0.73 at 24, illustrating how smaller fonts distort meaning due to cumulative recognition errors. Tajik shows smaller yet consistent gains, while Albanian and English maintain near-perfect BLEU scores, reaffirming their structural simplicity and visual clarity across all scales.

\vspace{-1em}
\vspace{-1em}

These results underscore font size as a crucial factor influencing OCR performance, especially for scripts with intricate structures like Urdu and Tajik. Targeted pre-processing techniques, such as adaptive scaling and resolution enhancement, are essential to address challenges at smaller sizes. Further, expanding annotated datasets and fine-tuning models for low-resource scripts can improve robustness, ensuring equitable OCR performance across diverse languages and reducing the reliance on pre-training biases favoring high-resource Latin scripts.

\vspace{-0.5em}
\vspace{-1em}


\subsection{Impact of Background Color}

\begin{figure}[h!]
    \centering
    \includegraphics[width=0.48\textwidth, height=0.4\textwidth, keepaspectratio]{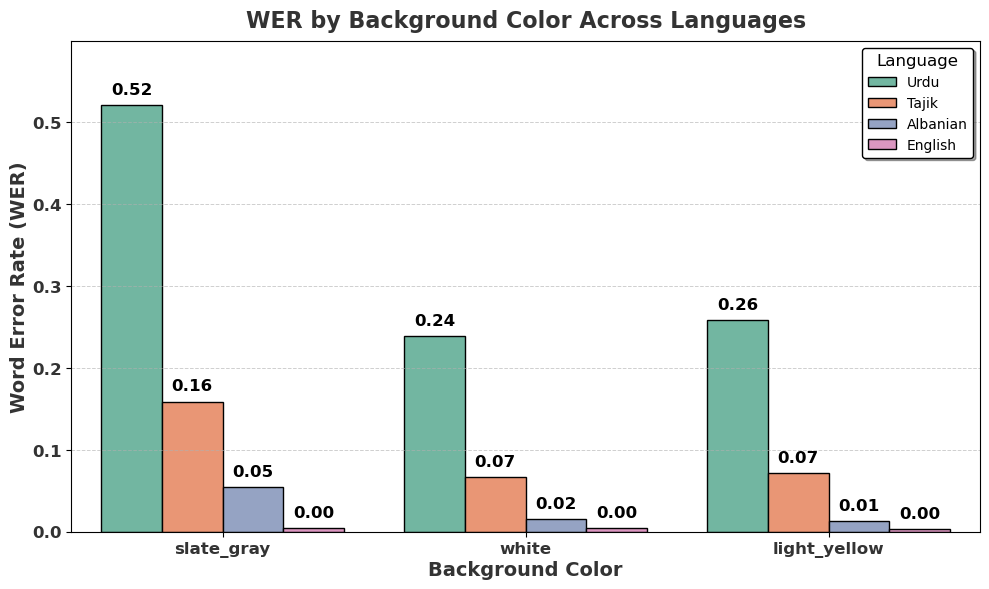}
    \caption{\small WER with Different Background Colors}
    \label{fig:wer_bg_color}
\end{figure}

Our analysis on background color highlights its varying influence on OCR performance across scripts. Slate gray’s low-contrast nature impairs the model’s ability to delineate character boundaries, a problem further exacerbated in scripts with dense diacritics and fused ligatures. This issue stems from OCR systems’ reliance on contrast-based edge detection, where perceptual noise reduces visual sharpness, leading to increased misalignment and character-level inaccuracies. This effect is shown in Figure~\ref{fig:wer_bg_color}, with the effect most pronounced in Urdu and moderately impactful in Tajik. For example, Urdu's WER spikes to 0.52, more than double its error on white (0.24) or light yellow (0.26). The intricate nature of Urdu’s ligatures and diacritics amplifies the challenge, as lower contrast diminishes visual clarity, making character boundaries harder to resolve. Tajik’s moderate degradation highlights the sensitivity of modified Cyrillic scripts to visual conditions. While simpler in structure compared to Urdu, the reduced contrast still introduces errors by complicating stroke recognition. This performance gap underscores the importance of visual clarity even for less intricate scripts when operating under non-ideal conditions. Also, The near-equivalent performance on white and light yellow backgrounds in the plot indicates the model’s robustness to slight variations in contrast.

\vspace{-0.8em}
\vspace{-0.7em}

\begin{figure}[h!]
    \centering
    \includegraphics[width=0.48\textwidth, height=0.4\textwidth, keepaspectratio]{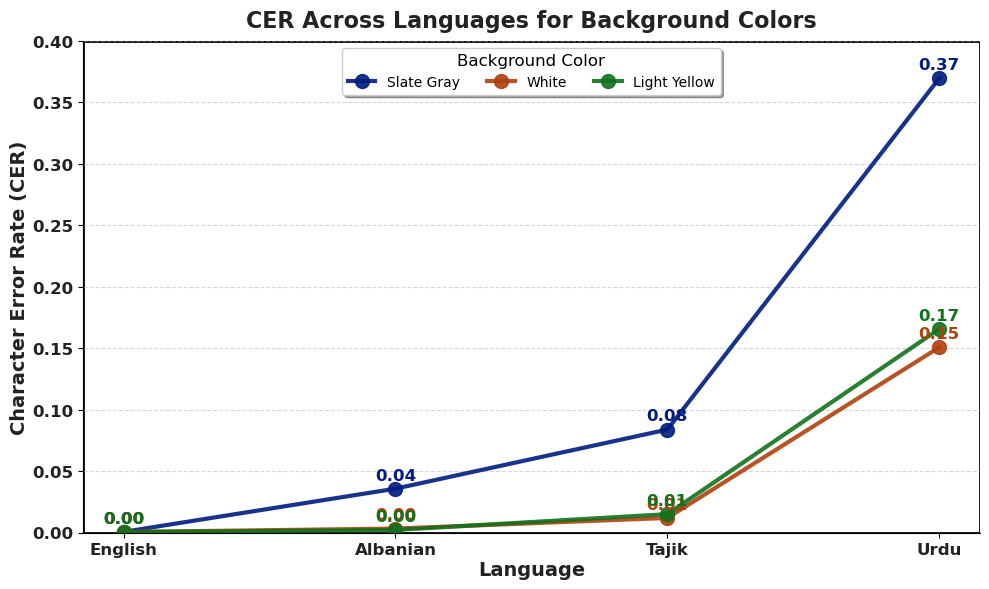}
    \caption{\small CER with Different Background Colors}
    \label{fig:cer_bg_color}
\end{figure}

\begin{figure}[h!]
    \centering
    \includegraphics[width=0.48\textwidth, height=0.4\textwidth, keepaspectratio]{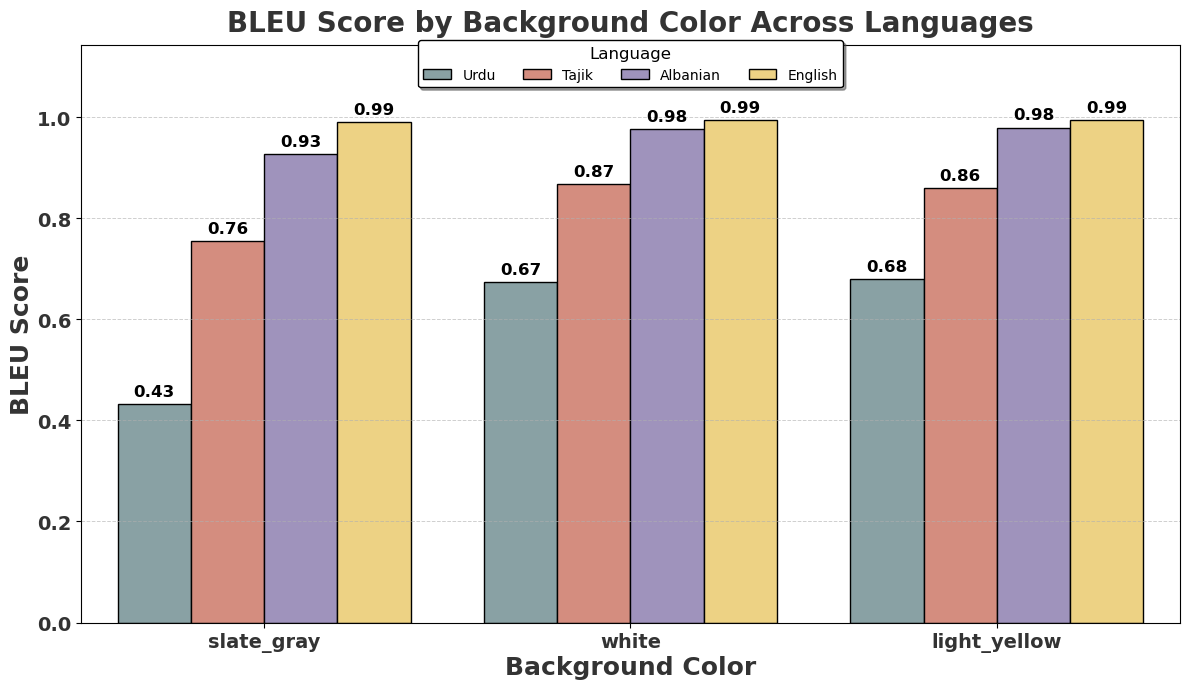}
    \caption{\small BLEU score with Different Background Colors}
    \label{fig:bleu_bg_color}
\end{figure}

In contrast, Albanian and English exhibit remarkable stability, with near-zero WER and CER (shown in Figure ~\ref{fig:cer_bg_color}) across all backgrounds. This robustness stems from their structural simplicity and alignment with GPT-4o’s pre-training, which predominantly involves Latin-based scripts. The minimal impact observed for these languages underscores the inherent advantage of high-resource scripts, which remain resilient to visual perturbations like contrast variations. As seen in Figure\ref{fig:bleu_bg_color}, the BLEU scores for Albanian and English remain consistently high, reinforcing that OCR systems perform exceptionally well for languages whose scripts align with model pre-training biases.

\vspace{-0.8em}
\vspace{-0.8em}
The degradation observed under slate gray backgrounds mimics real-world challenges encountered in digitizing historical documents, faded manuscripts, or low-quality scans where contrast is severely diminished. Developing OCR systems that can handle such conditions is crucial for enabling the preservation and accessibility of cultural and linguistic heritage. Future OCR systems must incorporate contrast-aware pre-processing techniques such as adaptive thresholding and contrast normalization to compensate for low-contrast degradation. Additionally, fine-tuning models on datasets augmented with varying contrast levels, particularly for underrepresented scripts, will ensure greater resilience. This approach would address real-world challenges where documents often suffer from faded text and poor print quality, ultimately promoting inclusivity for low-resource scripts.

\vspace{-1em}
\subsection{Impact of Gaussian Blur}

\begin{figure}[h!]
    \centering
    \includegraphics[width=0.48\textwidth, height=0.4\textwidth, keepaspectratio]{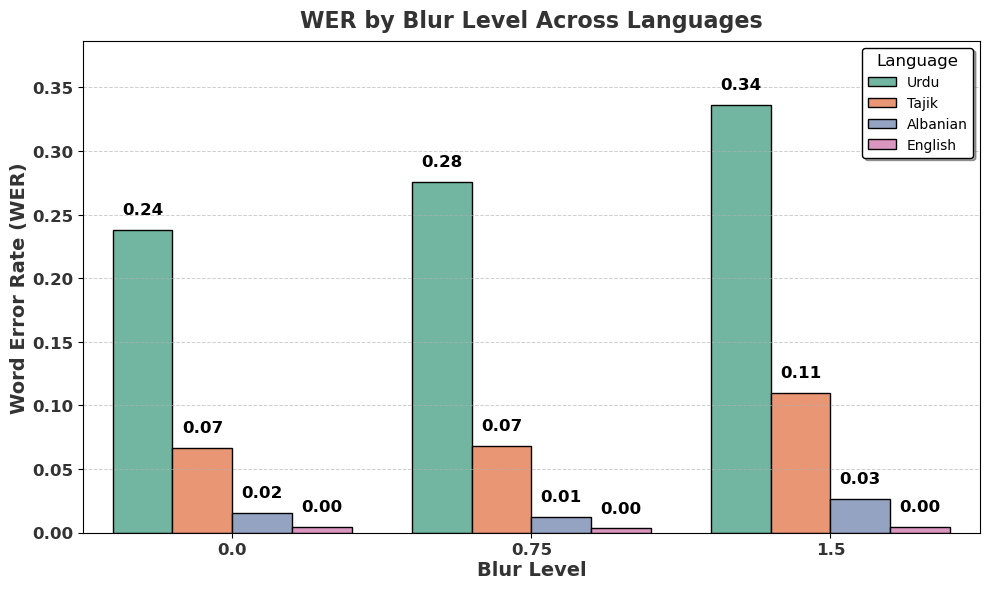}
    \caption{\small WER with Increasing Gaussian Blur}
    \label{fig:wer_blur}
\end{figure}

Gaussian blur introduces a clear degradation in OCR performance across all tested languages, with the magnitude of impact varying based on script complexity. As shown in Figure~\ref{fig:wer_blur}, WER increases progressively with higher blur levels, underscoring the challenges of maintaining character clarity when visual details are distorted.

\vspace{-1em}

\begin{figure}[h!]
    \centering
    \includegraphics[width=0.48\textwidth, height=0.4\textwidth, keepaspectratio]{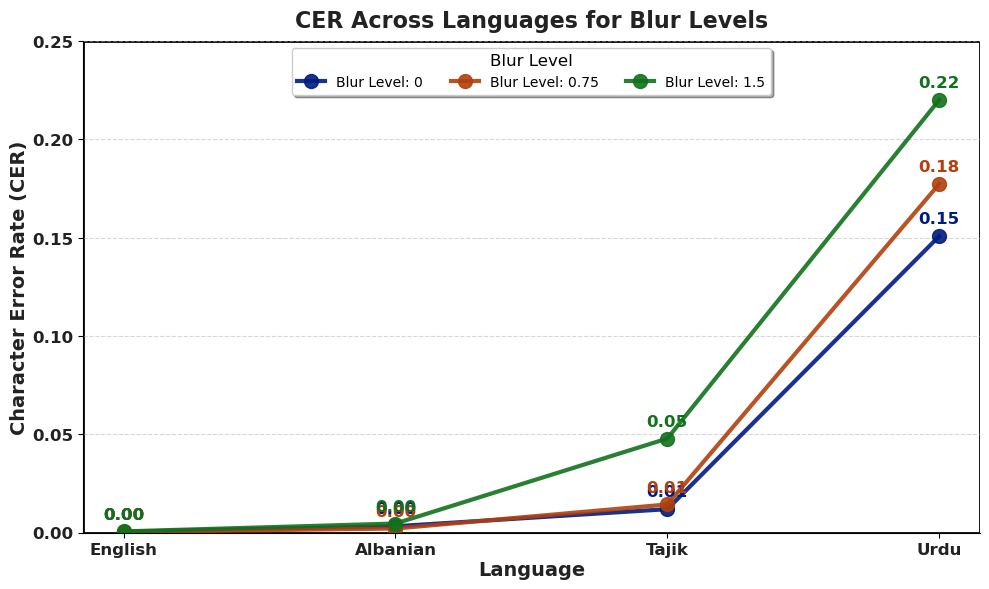}
    \caption{\small CER with Increasing Gaussian Blur}
    \label{fig:cer_blur}
\end{figure}

Blur has a pronounced effect on WER for Urdu, rising from 0.24 to 0.34 as blur levels increase from 0 to 1.5. This pattern is further reflected in Figure~\ref{fig:cer_blur}, where Character Error Rate (CER) climbs to 0.22. These results indicate that Urdu’s diacritics, dense ligatures, and non-linear script arrangement amplify the challenges under visual distortion, as blurred boundaries disrupt accurate recognition. However, Tajik exhibits a slower, more gradual rise in WER and CER, reaching 0.11 and 0.09, respectively. This resilience highlights the relatively linear structure of Tajik’s Cyrillic script, which mitigates the impact of blurring but does not entirely overcome its effects.
\vspace{-0.5em}
\vspace{-1em}

\begin{figure}[h!]
    \centering
    \includegraphics[width=0.48\textwidth, height=0.4\textwidth, keepaspectratio]{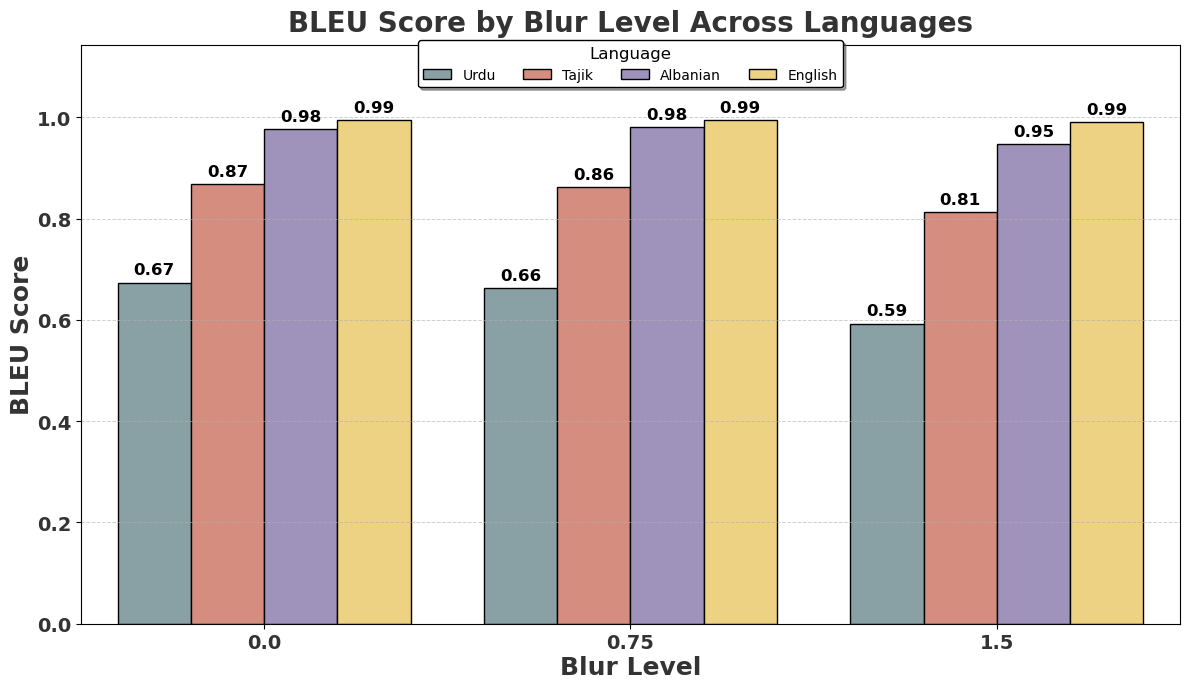}
    \caption{\small BLEU score with Increasing Gaussian Blur}
    \label{fig:bleu_blur}
\end{figure}

In contrast, Albanian and English demonstrate exceptional stability under increasing blur, with WER and CER remaining negligible across all blur levels. This robustness, as illustrated in Figure~\ref{fig:bleu_blur}, is further validated by consistently high BLEU scores nearing 0.99. The simplicity and clear spacing of Latin-based scripts, coupled with GPT-4o’s strong pretraining exposure to these languages, allow OCR systems to retain performance even when fine visual details are obscured. The minimal impact on Albanian also highlights how structural similarities with English confer additional advantages, despite its low-resource status.

\vspace{-0.5em}
\vspace{-1em}

These results emphasize that Gaussian blur disproportionately affects visually intricate scripts while exerting minimal influence on simpler, well-represented languages. Future OCR systems can address this disparity by incorporating blur-specific enhancements such as edge detection filters and deblurring techniques into preprocessing pipelines. Moreover, synthetic data augmentation with varying degrees of blur during model training could improve robustness across all scripts. By ensuring that models adapt to real-world visual distortions, such as defocused scans or motion blur, OCR systems can deliver equitable performance for both low-resource and structurally complex scripts, bridging existing gaps in accuracy.
\section{Limitations}

A significant limitation of this study was the labor-intensive nature of dataset creation. The manual formatting of text into images, combined with the need to ensure linguistic and visual consistency across four languages, imposed substantial time and effort constraints. This meticulous process limited the dataset size, preventing the inclusion of additional languages and larger datasets, which could have provided a more comprehensive evaluation of LLM-based OCR systems for low-resource scripts.

Another critical challenge was the prohibitively high cost of GPT-4o inference, which significantly restricted the scale of experimentation. The financial burden made it impractical to explore a broader range of variables or process larger datasets, underscoring the accessibility barriers posed by advanced language models. These constraints highlight the need for more scalable and cost-efficient approaches to facilitate robust OCR research across diverse linguistic landscapes.

\section{Future Work}

Building on the insights from this study, future research should aim to scale OCR evaluations across a wider array of low-resource languages to uncover broader linguistic patterns and challenges. Expanding datasets to include handwriting recognition and scene text scenarios, alongside variations like text orientation, noise artifacts, and occlusions, would provide a more comprehensive understanding of real-world OCR applications. These directions can help adapt LLM-based OCR to increasingly complex and diverse use cases.

Moreover, addressing the prohibitive costs associated with models like GPT-4o, future work could explore the development of open-source and lightweight alternatives tailored for low-resource settings. Fine-tuning pre-trained models on specific languages or leveraging model distillation techniques may improve accessibility without compromising performance. Establishing standardized, publicly available datasets and benchmarks for multilingual OCR would further democratize research and accelerate global collaboration, ensuring underserved communities benefit equitably from advancements in text digitization.

\vspace{-0.5em}
\section{Conclusion}
This study highlights the challenges and potential of LLM-based OCR systems, such as GPT-4o, in handling low-resource scripts like Urdu, Tajik, and Albanian across diverse linguistic and visual conditions. Our findings reveal that script complexity, low contrast, and visual distortions disproportionately degrade OCR accuracy for intricate scripts, while simpler Latin-based scripts like English and Albanian exhibit remarkable resilience. These disparities underscore the urgent need for script-aware preprocessing techniques, enriched annotated datasets, and fine-tuning strategies tailored for underrepresented scripts. By addressing these limitations and fostering inclusive OCR benchmarks, future systems can bridge existing gaps, ensuring equitable digitization of textual heritage and broadening accessibility for low-resource languages in an increasingly digital world.
\vspace{-0.5em}
\section{Acknowledgments}

We sincerely thank our colleagues Danish Humair and Shayan Ali Hassan for their valuable guidance and encouragement during the course of this research. Their support provided us with meaningful insights that helped refine our research approach.

\section{Supplementary Materials}
All resources developed and used in this study, including the curated dataset, the complete source code for experimental setups, and the detailed results, are available in our GitHub repository accessible \href{https://github.com/abdullahsohaill/CS6303-ResearchProject}{here}. This repository aims to promote transparency, reproducibility, and further research in LLM-based OCR for low-resource scripts.

\bibliographystyle{ACM-Reference-Format}
\bibliography{references}

\clearpage
\onecolumn

\appendix
\section{Appendix}

This appendix presents representative examples from the dataset used in our analysis, showcasing the variations across all tested dimensions: word count, font size, background color, and blur level. For brevity, we have included samples specifically for English, as the dataset curation process was standardized across all languages.

The complete dataset, including examples for other languages, is available in the accompanying GitHub repository for further reference. The selected images illustrate the range of dimensions under which the OCR performance was evaluated.

\subsection{Baseline Image}
\begin{figure}[h!]
    \centering
    \fbox{
        \includegraphics[width=0.6\textwidth, height=0.4\textwidth, keepaspectratio]{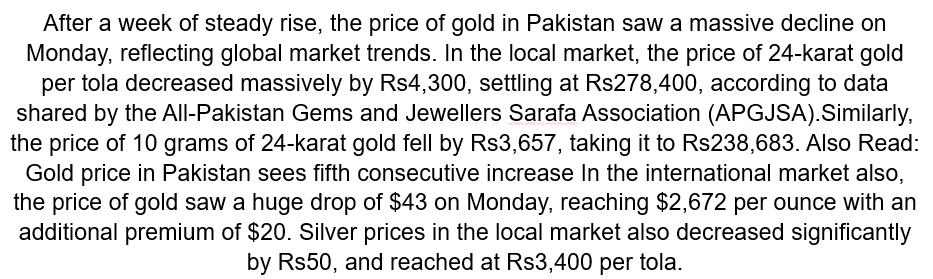}
    }
    \caption{\small Baseline image with word count 110-130, font size 18, white background, and no blur.}
    \label{fig:baseline}
\end{figure}

\subsection{Analysing the Impact of Word Count}
\begin{figure}[h!]
    \centering
    \fbox{
        \includegraphics[width=0.6\textwidth, height=0.4\textwidth, keepaspectratio]{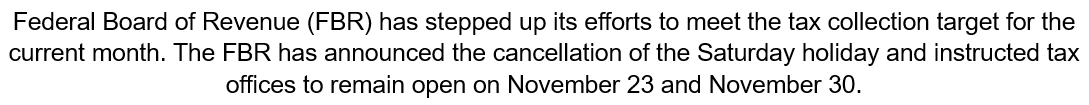}
    }
    \caption{\small Image with word count range 40-60.}
    \label{fig:wordcount_40-60}
\end{figure}

\begin{figure}[h!]
    \centering
    \fbox{
        \includegraphics[width=0.6\textwidth, height=0.4\textwidth, keepaspectratio]{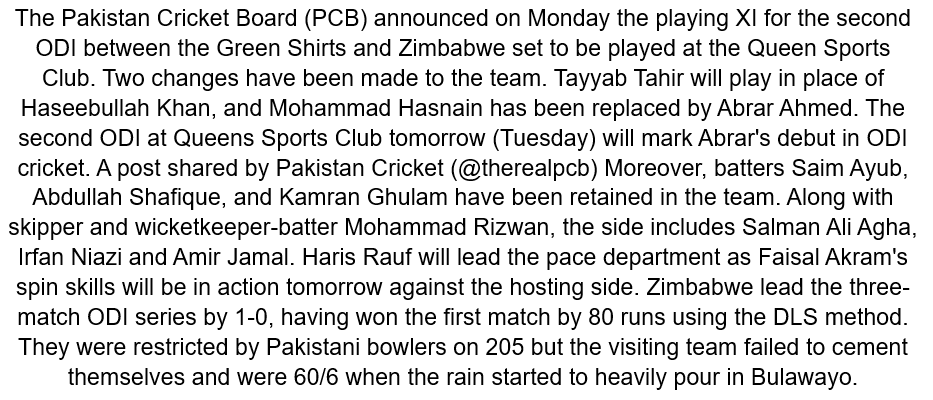}
    }
    \caption{\small Image with word count range 180-200.}
    \label{fig:wordcount_180-200}
\end{figure}

\subsection{Analysing the Impact of Font Size}
\begin{figure}[h!]
    \centering
    \fbox{
        \includegraphics[width=0.6\textwidth, height=0.4\textwidth, keepaspectratio]{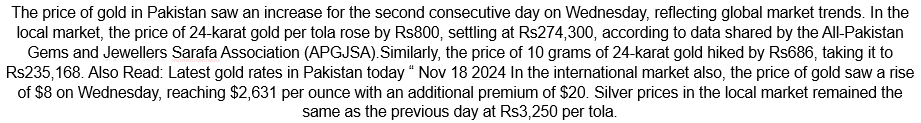}
    }
    \caption{\small Image with font size 12}
    \label{fig:fontsize_12}
\end{figure}

\clearpage

\begin{figure}[h!]
    \centering
    \fbox{
        \includegraphics[width=0.6\textwidth, height=0.4\textwidth, keepaspectratio]{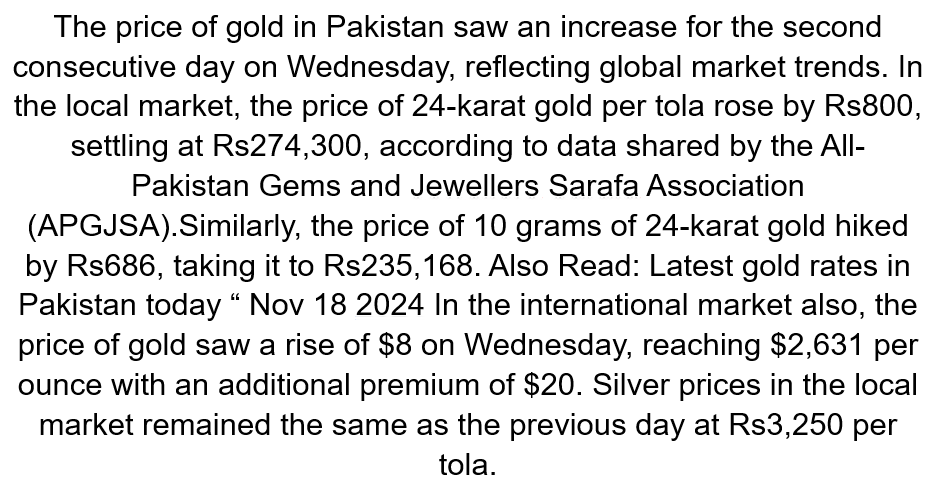}
    }
    \caption{\small Image with font size 24}
    \label{fig:fontsize_24}
\end{figure}

\subsection{Analysing the Impact of Background Color}
\begin{figure}[h!]
    \centering
    \fbox{
        \includegraphics[width=0.6\textwidth, height=0.4\textwidth, keepaspectratio]{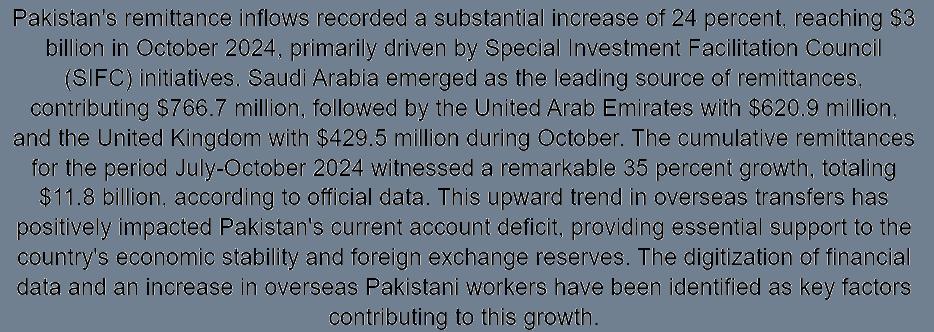}
    }
    \caption{\small Image with slate gray background}
    \label{fig:bg_slategray}
\end{figure}

\begin{figure}[h!]
    \centering
    \fbox{
        \includegraphics[width=0.6\textwidth, height=0.4\textwidth, keepaspectratio]{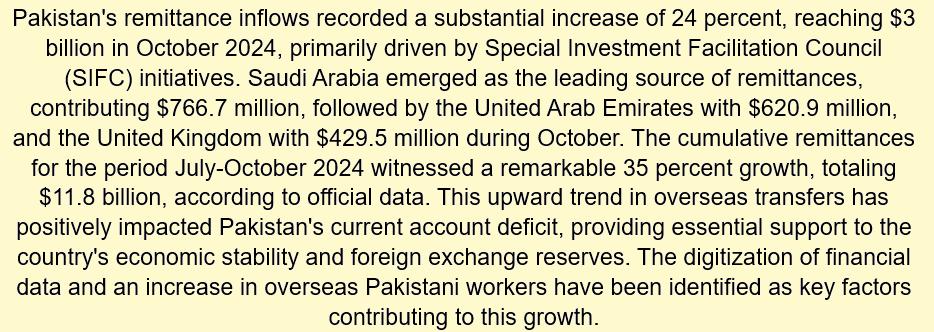}
    }
    \caption{\small Image with light yellow background}
    \label{fig:bg_lightyellow}
\end{figure}

\clearpage

\subsection{Analysing the Impact of Gaussian Blur}
\begin{figure}[h!]
    \centering
    \fbox{
        \includegraphics[width=0.6\textwidth, height=0.4\textwidth, keepaspectratio]{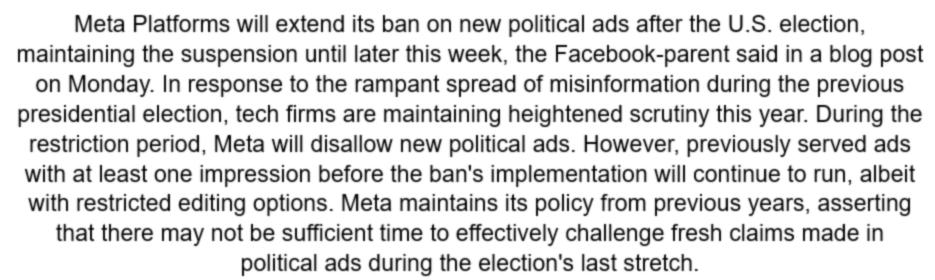}
    }
    \caption{\small Image with Gaussian blur level 0.75}
    \label{fig:blur_075}
\end{figure}

\begin{figure}[h!]
    \centering
    \fbox{
        \includegraphics[width=0.6\textwidth, height=0.4\textwidth, keepaspectratio]{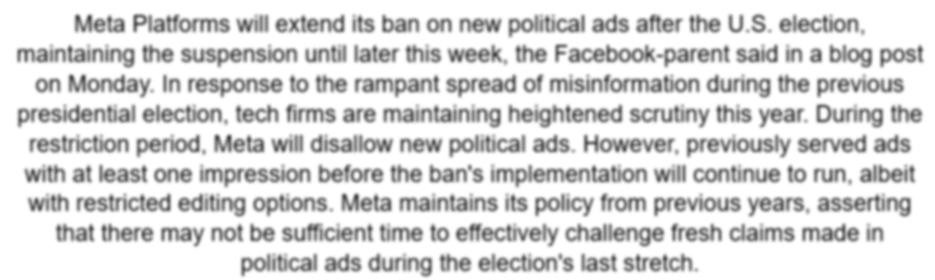}
    }
    \caption{\small Image with Gaussian blur level 1.5}
    \label{fig:blur_15}
\end{figure}

\end{document}